\documentclass[runningheads]{llncs}
\usepackage{etoolbox} 
\makeatletter
\patchcmd{\ps@headings}{\rlap{\thepage}}{}{}{}
\patchcmd{\ps@headings}{\llap{\thepage}}{}{}{}
\makeatother
\usepackage{graphicx}

\usepackage{amsmath}
\usepackage{amssymb}
\usepackage{caption}
\usepackage{mathtools, nccmath}
\usepackage{times}
\usepackage{epsfig}

\usepackage[pagebackref=false,breaklinks=true,letterpaper=true,colorlinks,bookmarks=false]{hyperref}
\usepackage{url}
\usepackage[nameinlink,capitalise]{cleveref}
\usepackage[utf8]{inputenc}
\usepackage{subcaption}

\DeclarePairedDelimiter{\round}\lfloor\rceil

\begin{document}
\title{Generating Natural Adversarial Hyperspectral examples with a modified Wasserstein GAN}

\author{Jean-Christophe Burnel \inst{1}\thanks{\mailsa}, 
Kilian Fatras\inst{1}, 
Nicolas Courty\inst{1}}

\urldef{\mailsa}\path|Contact author: jean-christophe.burnel@irisa.fr|
\authorrunning{}

\institute{University of South Brittany, CNRS, IRISA, UMR 6074, France}
\maketitle              
\begin{abstract}

Adversarial examples are a hot topic due to their abilities to fool a classifier's prediction. There are two strategies to create such examples, one uses the attacked classifier's gradients, while the other only requires access to the classifier's prediction. This is particularly appealing when the classifier is not full known (black box model). In this paper, we present a new method which is able to generate natural adversarial examples from the true data following the second paradigm. Based on Generative Adversarial Networks (GANs) \cite{goodfellow2014generative}, it reweights the true data empirical distribution to encourage the classifier to generate adversarial examples. We provide a proof of concept of our method by generating adversarial hyperspectral signatures on a remote sensing dataset. 
\end{abstract}
\section{Introduction}
 One of the weakness of DNNs is the so called adversarial example sensibility. Adversarial examples are samples which are not classified in the correct class. These examples can occur even for very accurate classifiers and can be problematic for their real-world applications, especially when security is an issue. Therefore one may want to have access to examples which can fool a classifier in order to measure how robust is the DNN.

Adversarial examples have received a lot of attention from the machine learning community. There are two main strategies to generate such examples. The first strategy is to use the gradients from classifiers. The generated adversarial examples are images drawn from the true distribution where we add a small perturbation following the gradients. The intuition is that the classifier will change its prediction because the gradients are the steepest classifier variation directions. This is the main idea of the Fast Gradient Sign Method (FGSM) algorithm \cite{goodfellow2015adv}. One problem from this approach is that one needs to know the full architecture to compute the classifier's gradients. The second paradigm does not require access to gradients. It looks for adversarial examples only by having access to the classifier's prediction. Our new method follows the second strategy as we detailed later.

 To generate adversarial examples without requiring to classifier's gradients, \cite{zhao2018generating} uses a GAN and an invertor. The generator is a function which goes from the latent space to the true data space while the invertor is an inverse function. From the true data, they use the invertor to go to the latent space and look for adversarial examples. They add small perturbations to the latent vector $z^*$ which represents a data $x^*$ until they find an adversarial example. This method allows them to have very realistic adversarial examples without using the classifier's gradients.
In \cite{song2018constructing} authors use an AC-GAN to model the data distribution with the assumption that an ideal model could generate all the set of legitimate data. With such a model they search in the latent space for all the adversarial examples. Their search is done by minimizing the confidence score of a classifier while having the auxiliary classifier still predicting the correct class. So the adversarial examples look like true images and are adversarial for the attacked classifier.

We propose a different approach to generate natural adversarial examples from those methods. Like them, our method only needs to know the output of a given classifier, seen as it a "black box" classifier. From a re-weighted distribution of the true data based on a classifier's predictions, a generator is learnt to generate adversarial examples. The idea is to create a map between the latent space and the set of adversarial images which are present in the true data.

The paper is structured as follows: after a brief introduction on GAN, we present our \textbf{main contributions} in section 3. We will detail our modified loss function for the WGAN architecture which is specialised in adversarial examples generation. In section 4, we will test our method on hyperspectral data from the Data Fusion Contest 2018 \cite{grss_dfc_2018}, by generating data which are misclassified by a classifier.

\section{Proposed methods}
{\bfseries Gerative Adversarial Networks.} The principle of a GAN is to learn a generator of realistic data. Intuitively, it tries to minimize the distance between the distributions of generated data and target data. This notion of distance between distribution is critical and has led to several variants of GAN. Formally, the generator takes $z \sim \mathbb{P}_Z$, as input to generate data $G(z) \sim \mathbb{P}_G$ and then distribution their distance to real data distribution $\mathbb{P}_r$. $\mathbb{P}_Z$ is usually a gaussian or a uniform distribution. GANs improve the generated data quality by minimizing the distance between the distributions $\mathbb{P}_G$ and $P_r$. However as we do not know the true distributions $\mathbb{P}_G$ and $\mathbb{P}_r$, we only consider their empirical distributions from the available true data and the generated ones. Wasserstein GAN \cite{arjovsky2017wasserstein} is a GAN variant which uses the Wasserstein distance to minimize the distance between distributions. Thanks to the Kantorovich-Rubinstein duality \cite{peyreOT} for calculating this Wasserstein distance, the objective function of WGAN is :
    \[\min_{\theta}\max_{\phi} \mathbb{E}_{x \sim \mathbb{P}_r}[\mathcal{D}_\phi(x)] -  
    \mathbb{E}_{z \sim \mathbb{P}_z}[\mathcal{D}_\phi(G_\theta(z))]
    \]
where $\mathcal{D}$ is within the set of 1-Lipschitz functions parameterized by $\theta$. Analogous to GANs, we still call \(\mathcal{D}\) the "discriminator" although it is actually a real-valued function. In order to respect the 1-Lipschitz constraint, \cite{arjovsky2017wasserstein} used a weight clipping trick for the discriminator's weights during the optimization procedure. However this practice is far from being optimal and might lead to unstability during optimization and poor minima. Another method \cite{petzka2017regularization} involves a gradient penalty and enforces the gradient to be less or equal to one. This is the variant of WGAN that we will consider in the rest of the paper:
    \[\min_{\theta}\max_{\phi} \mathbb{E}_{x \sim \mathbb{P}_r}[\mathcal{D}_\phi(x)] -  
    \mathbb{E}_{z \sim \mathbb{P}_z}[\mathcal{D}_\phi(G_\theta(z))] + 
    \mathbb{E}_{\hat{x} \sim \mathbb{P}_{\hat{x}}}[ (\max \{0, \| \nabla \mathcal{D}_\phi(\hat{x}) \|_2 - 1 \})^2] 
    \]
where \(\mathbb{P}_{\hat{x}}\) is the distribution of samples along the straight lines between a pair of points from \(\mathbb{P}_r \) and \(\mathbb{P}_G \).
\newline

{\bfseries Generating adversarial data.} For a fixed trained classifier on the true data, we aim to find a way to generate adversarial data with GANs. Instead of encouraging the generator to generate data close to the true data, we will encourage it to generate adversarial data which follows the true data distribution \(\mathbb{P}_r \). Intuitively, the generator will be a map between the latent space and the adversarial data from the true distribution. We opted for a modification of the loss function of WGAN by considering a reweighting strategy of the empirical true distribution. Indeed, The generator generates data similar to the true data because of the true data empirical distribution which is $\frac{1}{N} \sum_{i=1}^n \delta_{x_i}$. Our main idea is to reweight a data weight according to the classifier's prediction. If a data is misclassified, the resulting weight will be bigger than a correctly classified data. The resulting distribution must have the following form: $\sum_{i=1}^n p_i \delta_{x_i}$, where: $\sum_{i=1}^n p_i = 1$. Now we will describe and review the impact of different reweighting methods:
\newline


\par
\begin{figure*}[t!p]
    \centering
    \begin{subfigure}[t]{0.32\textwidth}
      \centering
        \includegraphics[width=0.9\textwidth]{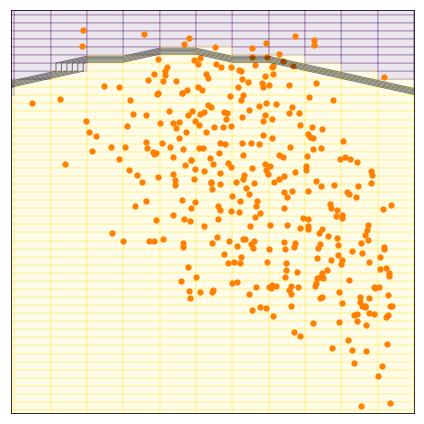}
      \vspace{-0.30cm}
      \captionof{figure}{}
      \label{fig:mean}
    \end{subfigure}%
    ~ 
    \begin{subfigure}[t]{0.32\textwidth}
      \centering
        \includegraphics[width=0.9\textwidth]{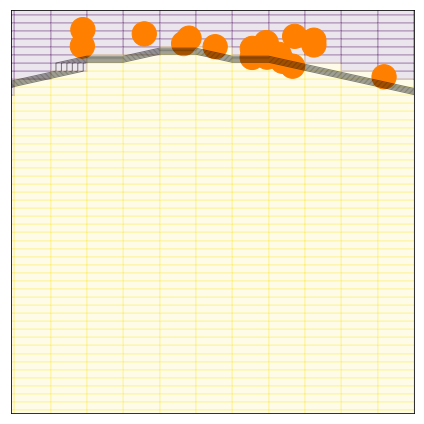}
      \vspace{-0.30cm}
      \captionof{figure}{}
      \label{fig:hard}
    \end{subfigure}%
    ~ 
    \begin{subfigure}[t]{0.32\textwidth}
      \centering
        \includegraphics[width=0.9\textwidth]{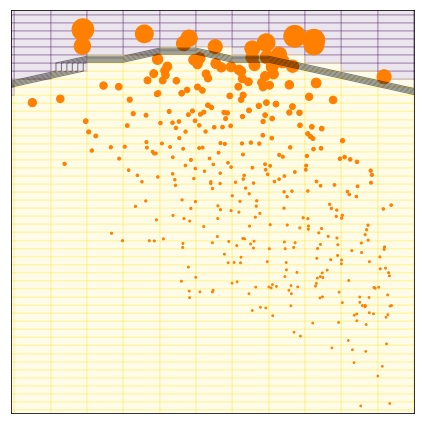}
      \vspace{-0.30cm}
      \captionof{figure}{}
      \label{fig:soft}
    \end{subfigure}%

    \begin{subfigure}[t]{0.32\textwidth}
      \centering
      \includegraphics[width=0.9\textwidth]{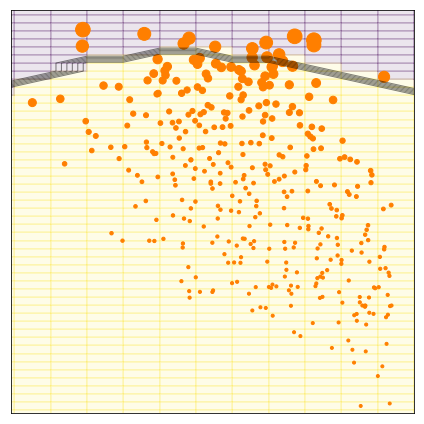}
      \vspace{-0.30cm}
      \captionof{figure}{\( w = 5\)}
      \label{fig:soft_5}
    \end{subfigure}%
    ~ 
    \begin{subfigure}[t]{0.32\textwidth}
      \centering
      \includegraphics[width=0.9\textwidth]{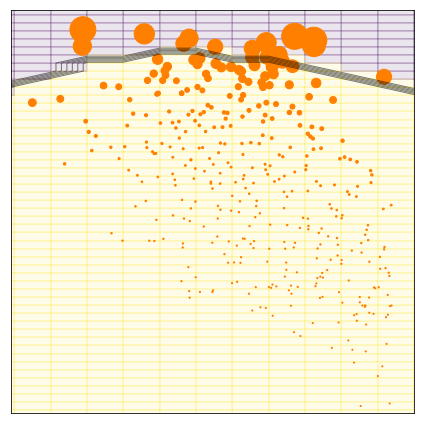}
      \vspace{-0.30cm}
      \captionof{figure}{\( w = 10\)}
      \label{fig:soft_10}
    \end{subfigure}
    ~ 
    \begin{subfigure}[t]{0.32\textwidth}
      \centering
      \includegraphics[width=0.9\textwidth]{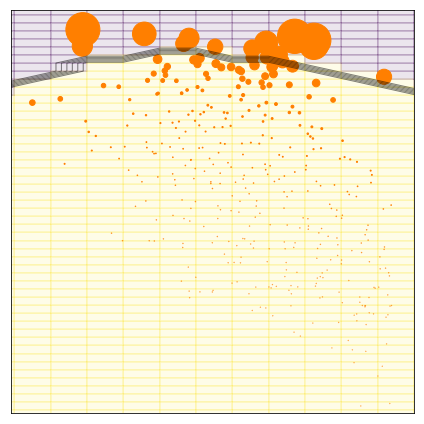}
      \vspace{-0.30cm}
      \captionof{figure}{\( w = 15\)}
      \label{fig:soft_15}
    \end{subfigure}
    \caption{Visualization of differents weighting methods}
    \vspace{-0.5cm}
\end{figure*}

 \textbf{Uniform weight} Here we are in the case of traditional WGAN, all data are given the same weight so there is no particular reason for the generator to produce adversarial examples, we can visualize it with figure \ref{fig:mean} where we have a distribution where all points got the same weight 1/N.

\par \textbf{Hard weighting} An intuitive way to generate adversarial example is to only consider misclassified data. In figure \ref{fig:hard}, we see the data given the following weighting : \(\round{1-c(x)} * x \), where \( c(x) \) is the classifier output. The downside of this weighting strategy is that we are left with few data for very accurate classifier. Unfortunately, GANs are data hungry and then this method is not efficient to generate adversarial examples.

\par \textbf{Soft weighting} Another weighting strategy is to use the prediction vector from the classifier. This can be done through a softmax function, defined as:
\begin{equation} %
\text{softmax}(x, w) \triangleq  \frac{\exp (w*[x_j - x_{max}])}{ \sum\limits_{i=0}^K \exp (w*[x_i - x_{max}])}  
\end{equation}
where \( x \) is our prediction vector and \( w \) is a temperature coefficient that controls the entropy of the rsulting distribution. In the figure \ref{fig:soft} we can see an example of weighting using the softmax function with a temperature \( w = 8 \).

To demonstrate the relevance of the different reweighting methods, we use a toy dataset to illustrate our approach from Figures \ref{fig:mean} to \ref{fig:soft_15}, where we represent data from a certain class with points, the classifier's decision boundary with a line, so that points under the line are correctly classified and points over the line are misclassified. 


As mentioned before, an important hyperparameter of our softmax function is the temprature coefficient. Its value allows controling the weight for data with low prediction score. In the next figure, we use a factor of 5 for figure \ref{fig:soft_5} and we can see that while misclassified data are given bigger weights the major part of the points still have a decent amount of weight. In figure \ref{fig:soft_10} the weights for data too far from the classification line start fading due to their weights being low and finally in figure \ref{fig:soft_15} we are vastly using the points along the classification line.

%

\section{Experiments and Results}
Unlike traditional RGB images, hyperspectral imagery divides the color spectrum in multiple contiguous bands that can outreach the visible spectrum. In those images a pixel is a spectrum, and we can identify materials in a pixel by analysing it because each materials have its own spectral signature. To test our algorithm we first considered a pixel-based classification task on hyperspectral images. The classifier takes a spectrum as input and will output a probability vector of belonging to one particular material. The classifier we used is based on \cite{hu2015deep} and have an Overall Accuracy of 0.52 with $\kappa = 0.43$ on disjoint train/test, which seem coherent with the result observe in recent reviews \cite{8738045}. Our objective here is to generate natural adversarial spectra.

\begin{figure}[h]
\centering
\begin{minipage}{.325\textwidth}
  \centering
  \includegraphics[width=0.9\textwidth]{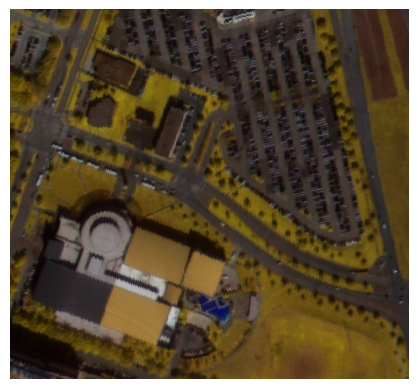}
  \captionof{figure}{False RGB}
  \label{fig:dfc_data}
\end{minipage}%
\begin{minipage}{.5\textwidth}
  \centering
  \includegraphics[width=0.9\textwidth]{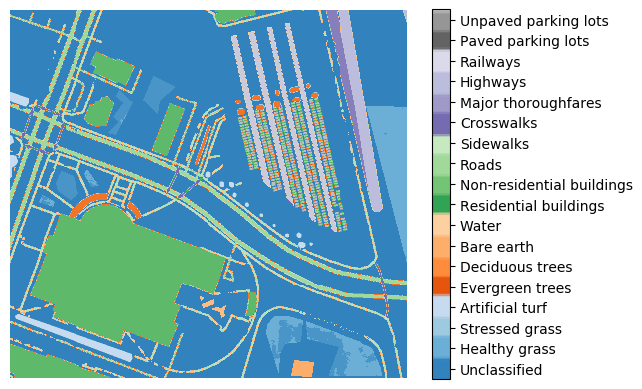}
  \hspace{-1.8cm}
  \captionof{figure}{Labels}
  \label{fig:dfc_labels}
\end{minipage}
\vspace{-0.5cm}
\end{figure}

{\bfseries Dataset.} We use the DFC2018 dataset \cite{grss_dfc_2018} which was acquired over the University of Houston campus and its neighbourhoods. It has hyperspectral data covering a 380-1050 nm spectral range with 48 bands at a 1-m GSD. Here the classes do not only cover urban classes (buildings, cars, railways, etc.) but it also has some vegetation classes (healthy or stressed grass, artificial turf, etc.). A view of the data is given with figure \ref{fig:dfc_data}, where we only take 3 of the 48 bands for visualization purpose, and the ground truth labels are shown with figure \ref{fig:dfc_labels} 

\begin{figure}
\includegraphics[width=\textwidth]{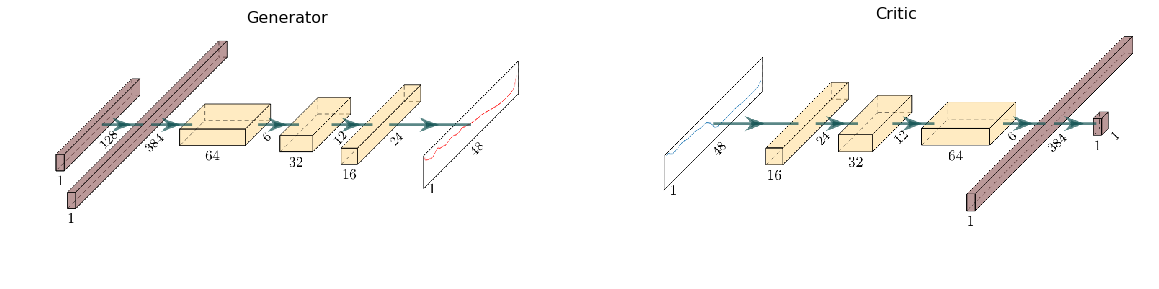}
\captionof{figure}{Architecture used for Generator and Critic}
\label{fig:arch}
\end{figure}

{\bfseries Experimentation.} For this experiment we want to generate spectra of healthy grass, we know that the classifier has an accuracy of 82\% for that particular class on the whole dataset. For the WGAN architecture we choose 1-dimensional convolutions for both the generator and the critic as it keeps coherence between close spectral bands. For the generator we have the followings layers (as seen in figure \ref{fig:arch}):\\ {\small\textbf{(Linear, 384)(Reshape, (6,64))(Conv1D, (12,32), 3)(Conv1D, (24,16), 3)(Conv1D, (48,1), 5)}},\\
and for the critic we have : \\
{\small\textbf{(Conv1D, (24,16), 5)(Conv1D, (12,32), 3)(Conv1D,(6,64), 3)(Reshape, (384))(Linear, 1)}}.\\ 
We use all the available data. In this experiment we choose to use \(w = 20\)

\begin{figure}
\centering
\includegraphics[width=\textwidth]{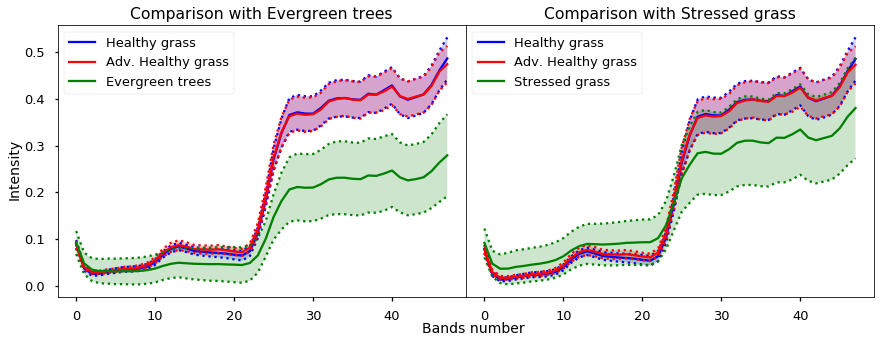}
\captionof{figure}{[Best viewed in color] Comparison with Evergreen trees and Stressed grass}
\label{fig:compare_both}
\end{figure}

\begin{figure}
\centering
\includegraphics[width=\textwidth]{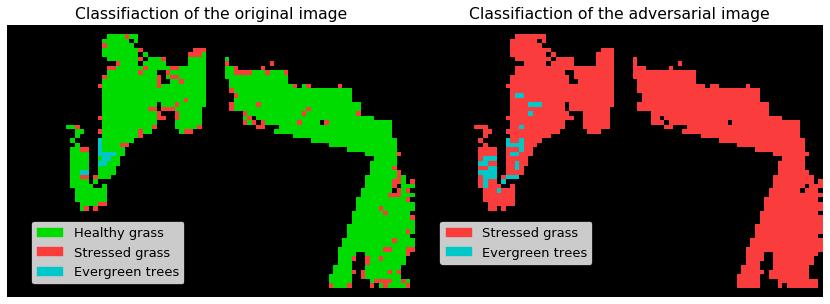}
\captionof{figure}{[Best viewed in color] Comparison of predictions}
\label{fig:compare_labels}
\end{figure}

{\bfseries Results.} 
We consider the healthy grass for target class, and we want to generate spectra similar to it but misclassified by the classifier. For this we used our ground truth to replace the pixels of the target class by the closest generated spectra.
To verify that our generated spectra belong to the target class, we choose to plot the statistics of the spectra, in plain the mean of the spectra and dotted the standard deviation. 
To interpret the results we choose to plot in the same figure the target class from the image, the adversarial examples generated and the two most predicted classes by our classifier. The laters are evergreen trees and stressed grass figure \ref{fig:compare_both}. In both cases we see that the adversarial spectra match the targeted class more than the predicted class.  With figure \ref{fig:compare_labels} we first show classification of healthy grass from the original image, here the classifier have 88.77\% accuracy, and the classification of the adversarial spectra.


\begin{figure}
\centering
\includegraphics[width=\textwidth]{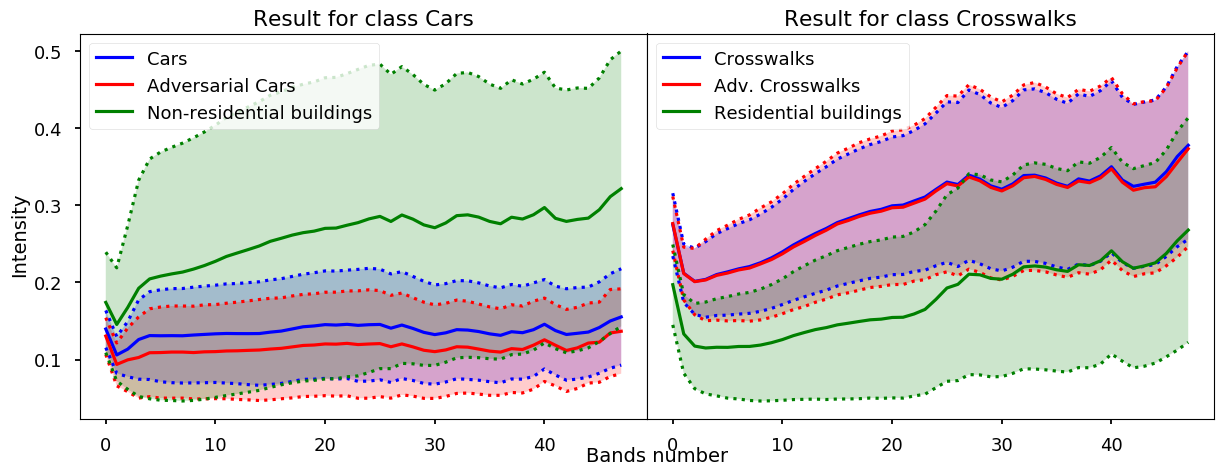}
\captionof{figure}{[Best viewed in color] Results for classes Car and Crosswalks}
\label{fig:compare_labels_18_12}
\end{figure}

To go further we show some results with two other classes with different accuracy.
We first choose Cars class where the attacked classifier have an accuracy of 0.39 and Crosswalks where the classifier have an accuracy of only 0.05. The results are visible with figure \ref{fig:compare_labels_18_12}, here the comparison is made with the most predicted class, and we can see that it works in both cases. Moreover we show in table \ref{tab:stats} that the more accurate our classifier is the less there are adversarial examples per batch. This can be explained by the fact that we do not have access to an infinite number of data so a more accurate classifier leads to less adversarial data to train the model.

\begin{table}[ht]
\centering
\caption{Succes rate of our method (over 10 runs)}
\label{tab:stats}
\begin{tabular}[t]{lccc}
\hline
&Mean &Std &Clf. Acc \\
\hline
Healthy Grass & 0.66 & 0.06 & 0.82\\
Car           & 0.81 & 0.05 & 0.39 \\
Crosswalks    & 0.99 & 0.009 & 0.05\\
\hline
\end{tabular}
\end{table}%

\begin{figure}
\centering
\includegraphics[width=\textwidth]{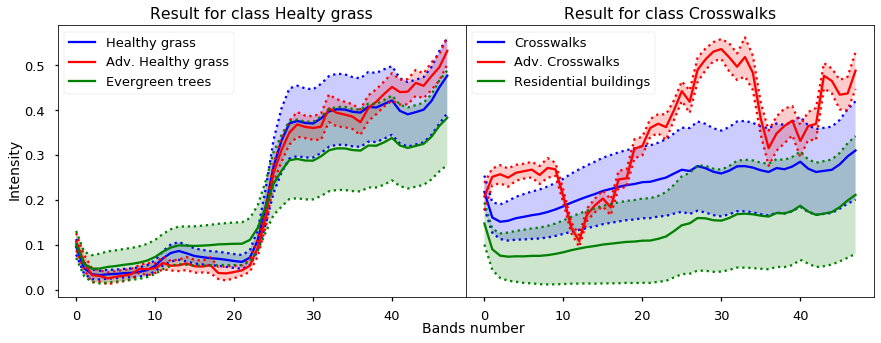}
\captionof{figure}{[Best viewed in color] Results of \cite{song2018constructing} for classes Healthy grass and Crosswalks}
\label{fig:compare_song}
\end{figure}

{\bfseries Qualitative comparison with state of the art.}
We now compare our results with \cite{song2018constructing}\footnote{following their online implementation : \url{https://github.com/ermongroup/generative_adversary}}. We adapted this method to the same considered architectures considered in this paper. We used untargeted attacks, which means that we want our spectra to be misclassified but we do not want it to be classified as a specific label. However, it is to note that their algorithm is capable of doing targeted attack. The results are visible with figure \ref{fig:compare_song}. We did not include class Car for it was not possible to obtain any adversarial example over 100 runs of 10 000 iterations. For the  Healthy grass class, the produced adversarial spectrum exhibits a lot of noise. For the class Crosswalks, it looks like their algorithm was not able to produce spectra close the real distribution, while our method worked in both cases. This somehow illustrates the benefits of our generation scheme. Further studies will try to confirm this view on a more extensive validation, involving practitioners from the field. 

\section{Conclusion and discussion}
In this work we propose a new method to generate natural adversarial examples for a pretrained classifier, using a modified loss function for the WGAN with a re-weighted distribution of our data based on the pretrained classifier's prediction. Our method departs from other generative networks in the sense that we do not generate adversarial samples from existing ones by addition of a small perturbation. We rather consider the case where we train a generator to produce adversarial samples. As an immediate consequence, our method can treat the initial pretrained classifier as a black-box,
which is a more realistic scenario.~\\
We applied our method to the generation of hyperspectral data and show that it is possible to generate spectra that fool a classifier while keeping a similar content. One possible extension of our work would be to adapt our method to generate other modalities frequently encountered in remote sensing, such as point cloud data, but also to generate entire portions of images rather than solely focusing on spectra (spatial-spectral cubes). Here again, apart from the complexity of the needed neural architectures for such tasks, we do not see any major issue that would prevent our method from working. Complementarily, extensive tests with photo-interprets will be conducted to further assess the qualities of our algorithm. 

%
%
%
%
\bibliography{biblio} 
\bibliographystyle{splncs04}
\end{document}